\documentclass[sigconf,natbib=True]{acmart}

\usepackage{booktabs}
\usepackage{adjustbox}
\usepackage{balance}
\AtBeginDocument{%
  \providecommand\BibTeX{{%
    \normalfont B\kern-0.5em{\scshape i\kern-0.25em b}\kern-0.8em\TeX}}}

\setcopyright{acmcopyright}

\copyrightyear{2024}
\acmYear{2024}
\setcopyright{acmlicensed}\acmConference[SIGIR '24] {Proceedings of the 47th International ACM SIGIR Conference on Research and Development in Information Retrieval}{July 14--18, 2024}{Washington, DC, USA}
\acmBooktitle{Proceedings of the 47th International ACM SIGIR Conference on Research and Development in Information Retrieval (SIGIR '24), July 14--18, 2024, Washington, DC, USA}
\acmDOI{10.1145/3626772.3657670}
\acmISBN{979-8-4007-0431-4/24/07}

\usepackage{multirow}
\usepackage{spverbatim}

\usepackage{svg}

\begin{document}

\title[Retrieval-Augmented ConvRec with Semi-Structured NL State Tracking]{Retrieval-Augmented Conversational Recommendation with Prompt-based Semi-Structured Natural Language State Tracking}


\author{Sara Kemper}
\authornote{These authors contributed equally to this research.}
\affiliation{
  \institution{University of Waterloo}
  \city{Waterloo}
  \state{Ontario}
  \country{Canada}
}

\author{Justin Cui}
\authornotemark[1]
\author{Kai Dicarlantonio}
\authornotemark[1]
\author{Kathy Lin}
\authornotemark[1]
\author{Danjie Tang}
\authornotemark[1]
\affiliation{
  \institution{University of Toronto}
  \city{Toronto}
  \state{Ontario}
  \country{Canada}
}

\author{Anton Korikov}
\email{anton.korikov@mail.utoronto.ca}
\author{Scott Sanner}
\affiliation{
  \institution{University of Toronto}
  \city{Toronto}
  \state{Ontario}
  \country{Canada}
}

\renewcommand{\shortauthors}{Sara Kemper et al.}

\begin{abstract}
Conversational recommendation (ConvRec) systems must understand rich and diverse natural language (NL) expressions of user preferences and intents, often communicated in an indirect manner (e.g., ``\textit{I'm watching my weight}'').  Such complex utterances make retrieving relevant items challenging, especially if only using often incomplete or out-of-date metadata. Fortunately, many domains feature rich item reviews that cover standard metadata categories \textit{and} offer complex opinions that might match a user's interests (e.g., ``\textit{classy joint for a date}'').  However, only recently have large language models (LLMs) let us unlock the commonsense connections between user preference utterances and complex language in user-generated reviews.  Further, LLMs enable novel paradigms for semi-structured dialogue state tracking, complex intent and preference understanding, \emph{and} generating recommendations, explanations, and question answers.  We thus introduce a novel technology \textit{RA-Rec}, a \textbf{R}etrieval-\textbf{A}ugmented, LLM-driven dialogue state tracking system for Conv\textbf{Rec}, showcased with a video,\footnotemark[1] open source GitHub repository,\footnotemark[2] and interactive Google Colab notebook.\footnotemark[3]
%
\end{abstract}

\ccsdesc[500]{Information systems~Recommender systems}
\ccsdesc[500]{Information systems~Personalization}
\ccsdesc[300]{Information systems~Language models}

\keywords{Conversational Recommendation, LLM, Dialogue State Tracking}

\maketitle

\section{Introduction} \label{sec:intro}
Effective conversational recommendation (ConvRec) systems need to understand rich and diverse natural language (NL) expressions of user preferences and intents, often communicated in an indirect or subtle manner~\cite{harper2017natural,louis2020indirect,lyu2021workflow,hosseini2023indirect}. For instance, a user who asks ``\textit{Do they have parking?}'' is both \textit{inquiring} and \textit{providing a preference} for available parking. 
Similarly, a user looking for a restaurant who states ``\textit{I'm watching my weight}'' is expressing a complex preference that requires commonsense reasoning and may not match any predefined restaurant metadata fields. 
Metadata is also often incomplete or out-of-date, 
making it challenging to connect NL requests to relevant item recommendations. This creates major limitations for traditional NL ConvRec systems 
that rely on mapping user intents and preferences to predefined metadata taxonomies~\cite{yan2017shopping, sun2018conversational,narducci2020investigation, jannach2021survey}.

\begin{figure*}[t!]
   \centering
   \includegraphics[width=0.9\textwidth]{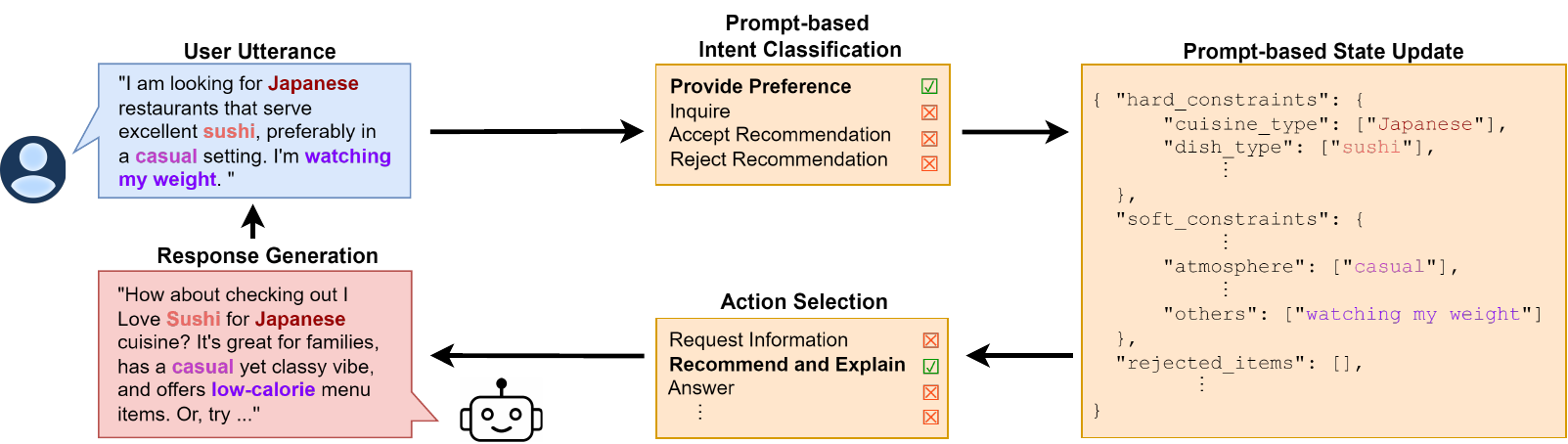}
   \caption{The \textit{RA-Rec} prompt-driven dialogue state tracking loop. LLM prompting is used for multi-label intent classification and for updating a JSON semi-structured NL state which tracks user preferences and other key dialogue elements. The state keys provide an easily configurable structure, while LLM-generated state values can capture nuanced NL expressions.}
   \label{fig:mainloop}
\end{figure*}

\begin{table*}[t!]
 \caption{User intents and system actions in \textit{RA-Rec}, which are a subset of the recommendation taxonomy of Lyu et al. \cite{lyu2021workflow}. 
 }
  \begin{adjustbox}{width=\textwidth, center}
  \begin{tabular}{llp{10cm}}
  \hline \hline 
  \noalign{\smallskip} 
  \multicolumn{3}{c}{\textbf{User Intents}} \\ \hline 
    \textbf{Intent} &  \textbf{Description} & \textbf{Examples} \\
  \midrule
    Provide Preference &  Provide or refine  preference for their desired item & ``I want a place with a very good scenic view.'' \\
    Inquire & Ask for more information about the recommended item(s) & ``What kind of menu do they offer?'', ``How do these options compare for price?'' \\
    Reject Recommendation & Reject a recommended item, either explicitly or implicitly & ``Probably too expensive, what else is there?'' \\
    Accept Recommendation & Accept a recommended item, either explicitly or implicitly & ``The first place looks good!'' \\
  \noalign{\smallskip}
       \hline \hline 
  \noalign{\smallskip} 
  \multicolumn{3}{c}{\textbf{System Actions}} \\ \hline 
    \textbf{Action} &  \textbf{Description} & \textbf{Examples} \\
  \midrule
    Request Information &  Request the user’s preferences towards item aspect(s) & ``Where are you located?'', ``What kind of cuisine are you looking for?'' \\
    Recommend and Explain & Recommend item(s) and explain how they match user preferences & ``How about trying Washoku Bistro for a comfortable and laid-back vibe while enjoying some delicious Japanese sushi?'' \\
    Answer & Respond to user inquiry about recommended item(s) & ``Yes, Tokyo Express has a parking lot.'' \\
    Respond to Rejection & Respond to user's rejection of recommended item(s) & ``I'm sorry that you did not like the recommendation. Is there anything else I can assist you with?'' \\
    Respond to Acceptance & Respond to user's acceptance of recommended item(s) & ``Great! If you need any more assistance, feel free to ask.'' \\
    Greeting & Greet the user. & ``Hello there! I am an Edmonton restaurant recommender. How can I help you?'' \\
       \hline
  \end{tabular}   
  \end{adjustbox}
 \label{tab: user intents WWW}
\end{table*}

Fortunately, many recommendation domains have an abundance of rich NL item reviews that not only refer to standard metadata categories but also offer more complex opinions and narratives that might match a user's interests, e.g. \textit{``The menu had lots of low-cal veggie options!''}. \ However, what we have lacked until recently with the advent of large language models (LLMs)~\cite{chatgpt,palm,sparks_of_agi} is the ability to unlock the commonsense reasoning connections between rich user preference utterances and expressive language in user-generated content such as NL reviews.  In addition to bridging this language expression and reasoning gap, LLMs also provide novel opportunities to control and facilitate a range of interactions in ConvRec dialogue, such as understanding user intents and preferences, \emph{and} generating recommendations, explanations, and answers to questions~\cite{deldjoo2024review}.

\begin{figure*}[h]
   \centering
   \includegraphics[width=0.95\textwidth]{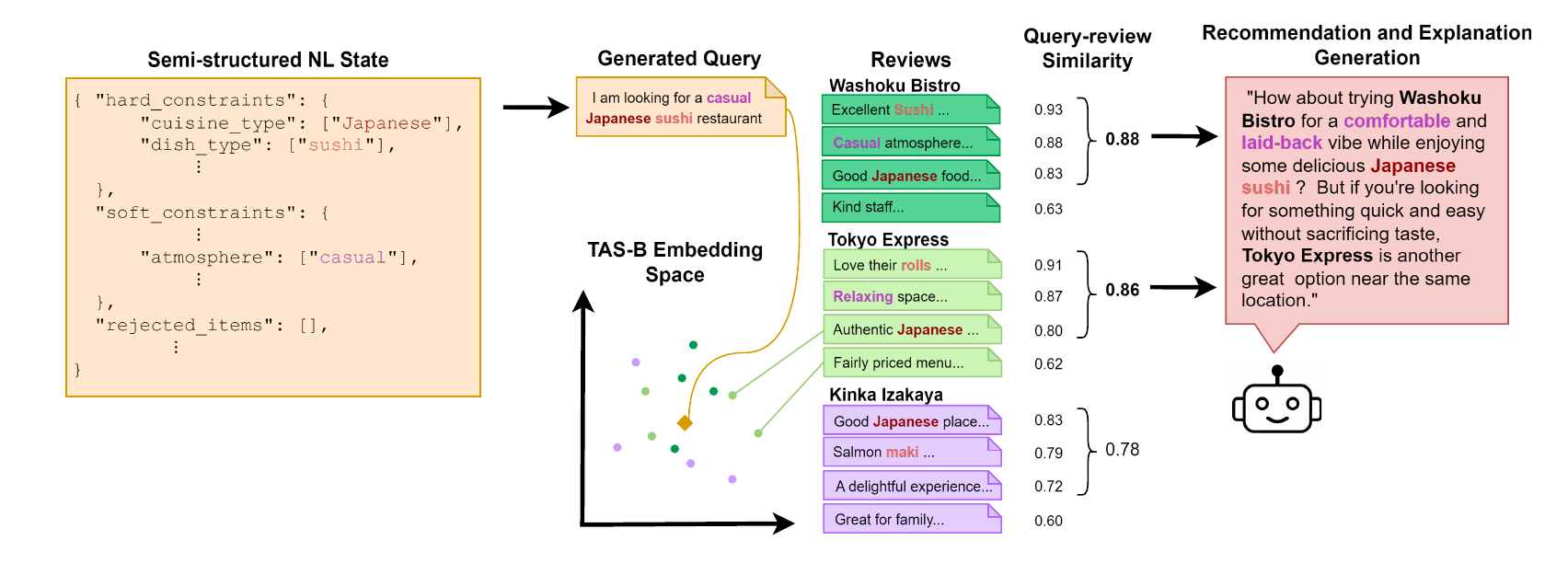}
   \caption{Retrieval-augmented recommendation and explanation using late fusion RIR. First, preferences in the dialogue state are used to generate a NL query. Then, query and review embeddings are scored using dot product similarity, and the top review scores for each item are averaged (fused) into an item score. Finally, the top items' most relevant reviews and metadata are used to generate a recommendation and explanation of how the item satisfies the preferences in the state.}
   \label{fig:late_fusion}
\end{figure*}

We thus introduce a novel open source demonstration technology \textit{RA-Rec}, a \textbf{R}etrieval-\textbf{A}ugmented, LLM-driven dialogue state tracking system for Conv\textbf{Rec}, making the following contributions:
\begin{itemize}
    \item We introduce prompt-driven ConvRec intent classification and state updating that captures nuanced NL expressions while maintaining domain-specific preference structure via a \emph{semi-structured} NL dialogue state (Sec.~\ref{sec:state update}).
    \item We extend recent work on \emph{reviewed-item retrieval}~\cite{abdollah2023self} to Conv-Rec dialogue, generating state-based queries, recommendations, explanations, and question answers (Figure~\ref{fig:late_fusion}).
    \item We demonstrate \textit{RA-Rec} for restaurant recommendation, including a video,\footnote{\href{https://www.youtube.com/watch?v=W8Y56UW2LTU}{https://www.youtube.com/watch?v=W8Y56UW2LTU}} a 
    well-documented open source GitHub repository under a \emph{permissive} MIT License,\footnote{\href{https://github.com/D3Mlab/llm-convrec}{https://github.com/D3Mlab/llm-convrec}} and an interactive Google Colab notebook that can run the system.\footnote{\href{https://apoj.short.gy/d3m-llm-convrec-demo}{https://apoj.short.gy/d3m-llm-convrec-demo}}
\end{itemize}

\section{Background and Related Work}
\subsection{Dialogue State Tracking} A standard Dialogue State Tracking (DST) loop \cite{williams2016dialog} has four steps: (1) intent understanding, (2) dialogue state updating, (3) action selection, and (4) response generation. A traditional state consists of keys and values, typically from a predefined set of labels such as \textit{``food: italian'', ``price: cheap'', ``area: east''}, that represent a most likely estimate of the participants' shared intentions and beliefs at a given turn \cite{cohen1987rational,williams2016dialog}. State tracking techniques generally map features extracted from user utterances to state labels, and include hand-crafted rules \cite{larsson2000information, bohus2003ravenclaw}, discriminative classifiers \cite{bohus2006k} and Bayesian networks \cite{paek2013conversation,williams2007partially}. While following the DST loop steps for modular dialogue control, our \textit{RA-Rec} system (Sec. \ref{sec:RARec}) extends traditional DST methods with LLM-driven state tracking to capture complex, NL expressions of preference and to facilitate state-based retrieval-augmented recommendation and question answering (QA).

\subsection{Reviewed Item Retrieval} \label{sec:RIR}
Aiming to unlock the expressive NL data available in reviews, Abdollah Pour et al. \cite{abdollah2023self} recently extended Neural IR \cite{reimers2019sentence} to an approach they call Reviewed Item Retrieval (RIR), where the key challenge lies in \textit{fusing} low-level information from multiple reviews to a higher item level~\cite{zhang2017design}. They demonstrate it is more effective to \textit{first} score individual reviews against a query and \textit{then} aggregate these scores to an item level (\textit{late fusion}), instead of summarizing reviews at an item level before query-scoring (\textit{early fusion}), since late fusion retains critical nuanced review information lost by early fusion.

In late fusion RIR, given a query and a set of reviews, a neural encoder maps each review and the query to respective embeddings.  A similarity function, such as the dot product, then computes a query-review similarity score. For each item, scores from the reviews with the highest query-review similarities are then averaged (fused) to give a query-item similarity score, and the top-scoring items are returned. As we will discuss in the next section, our \textit{RA-Rec} system adapts late fusion RIR to ConvRec by generating queries from a NL dialogue state and using review-based retrieval-augmented generation for recommendation and QA, as illustrated in Figure \ref{fig:late_fusion}.

\section{Retrieval Augmented Conversational Recommendation} \label{sec:RARec}

\begin{table*}[]
\caption{JSON keys in the \textit{RA-Rec} semi-structured NL dialogue state. Subkeys can be modified to easily facilate new domains. 
Bold subkeys indicate mandatory preferences the system will request information about if these preferences are not provided.} 
\centering
\footnotesize
\begin{tabular}{|l|p{6.5cm}|p{7cm}|}
\hline
\textbf{State Key} & \textbf{Description} & \textbf{Subkeys}  \\ \hline
hard\_constraints   & User preferences that must be satisfied  & \multirow{2}{7cm}{\textbf{location}, \textbf{cuisine\_type}, dish\_type, price\_range, atmosphere, dietary\_restrictions, wait\_times, type\_of\_meal, others} \\ \cline{1-2} 
soft\_constraints   & User preferences that are not required & \\ \hline
recommended\_items  & Previously recommended items  & -  \\ \hline
rejected\_items  & Previously rejected items & - \\ \hline          accepted\_items   &  Previously accepted items    & - \\ \hline        
\end{tabular}
\label{tab:state}
\end{table*}

To leverage both the modular structure of a traditional DST loop and the NL reasoning abilities of LLMs, we propose \textit{RA-Rec}, a modular, retrieval-augmented ConvRec system, illustrated in Figure \ref{fig:mainloop}. We employ a prompt-driven approach for intent classification and state updating, with the latter relying on a JSON format NL state that can be configured with domain-specific keys while capturing nuance through LLM-generated NL values. We then use this NL state to facilitate personalized, retrieval-augmented recommendation and QA utilizing item reviews and metadata. 

\subsection{Prompt-Driven Intent Classification}
After the user makes an utterance, LLM-prompting is used to determine whether the user expresses any of the four intents in Table \ref{tab: user intents WWW}, which are a subset of the recommendation dialogue intent taxonomy of Lyu et al. \cite{lyu2021workflow}. Table \ref{tab:prompts} outlines the prompts used in \textit{RA-Rec}, with full prompt templates available in the GitHub repository (see Sec. \ref{sec:intro}). We take a \emph{multi-label} intent classification approach to capture multiple intents that might be expressed in a single utterance --- for example, the utterance ``\textit{Does Washoku Bistro have parking?''} should be classified using both the intents \textit{``Inquire''} and \textit{``Provide preference''} because it expresses a preference towards available parking. A larger set of user intents can be facilitated by updating the system's prompts and initial state keys.   

\subsection{Semi-Structured NL Dialogue State Tracking} \label{sec:state update}
We store descriptions of user preferences and other important conversational elements such as rejected recommendations in a JSON state using the keys shown in Table \ref{tab:state} --- two state examples are in Figures \ref{fig:mainloop} and \ref{fig:late_fusion}. While the \textit{keys} provide structure, the state \textit{values} are typically LLM-generated from the latest utterances, allowing the state to represent complex NL expressions of preference such as \textit{``I'm watching my weight''} at a level of nuance and expressivity that would be impossible with predefined value sets. We thus refer to this state representation as a \textit{semi-structured} NL dialogue state.

\subsubsection{State Elements} Since the goal of \textit{RA-Rec} is recommendation, the most important components of the state maintain an up-to-date belief about user preferences, represented through hard (required) and soft (not required) constraints. In our restaurant recommendation demo, these constraints are represented with several domain-specific subkeys listed in Table \ref{tab:state}, as well as an ``\textit{others}'' subkey to capture any unspecified preference types. To adapt \textit{RA-Rec} to a new domain, these restaurant-specific subkeys can be replaced with domain-specific subkeys with little effort from a system designer. 

Other state elements include previously recommended, rejected, or accepted restaurants -- more elements could be easily added to handle a wider set of (domain-specific) user intents and system actions. Most state values are LLM-generated (prompts are summarized in Table \ref{tab:prompts}) and used downstream for action selection, recommendation, explanation, and QA, discussed next.

\begin{table*}[]
\caption{The main prompts used in \textit{RA-Rec} -- full templates can be found in the repository documentation (see Sec. \ref{sec:intro} link).} 
\centering
\footnotesize
\begin{tabular}{|p{2cm}|p{3cm}|p{11.5cm}|}
\hline
\textbf{Component} & \textbf{Prompt}  & \textbf{Description}   \\ \hline
Intent \mbox{Classification} & Classify Intent & Given a user utterance and description of an intent (e.g. inquire), identify whether the utterance expresses the intent.  \\ \hline
\multirow{2}{2cm}{State \mbox{Update}} & Update Constraints  & Given a user utterance and the previous hard and soft constraints, update the constraints.  \\ \cline{2-3}
 & Update Accepted/Rejected Item    & Given a user utterance with intent "Accept/Reject Recommendation", identify which item was accepted/rejected.    \\ \hline
\multirow{2}{2cm}{Recommendation \mbox{and} \mbox{Explanation}} & Generate \mbox{Recommendation} Query  & Given hard/soft constraints in the state, generate a NL query.   \\ \cline{2-3}
 & Explain Recommendations                             & Given the top retrieved items, their metadata, and their top reviews, explain how these recommended items match the hard/soft constraints.          \\ \hline
\multirow{4}{3cm}{QA} & Determine QA Knowledge Source & Given a user inquiry about recommended items and those items' metadata, identify which fields should be used to answer the inquiry, if any. If none, reviews will be used as the QA knowledge source. \\ \cline{2-3}
 & Answer Using Metadata                    & Given an inquiry and relevant metadata entries, generate an answer.  \\ \cline{2-3} & Generate QA Query & Given a user inquiry utterance, generate a NL query.  \\ \cline{2-3}
 & Answer Using \mbox{Reviews}   & Given an inquiry and retrieved reviews, generate an answer.    \\ \hline
\end{tabular}
\label{tab:prompts}
\end{table*}

\subsection{Action Selection}
The main system actions, summarized in Table \ref{tab: user intents WWW} are \textit{Request Information}, \textit{Recommend and Explain}, and \textit{Answer}. To understand our \textit{Request Information} implementation, consider a user asking for a restaurant recommendation without giving a location preference --- a recommendation may yield a restaurant in the wrong city!  To avoid such premature recommendations with insufficient context, we identify \textit{mandatory} preferences that the system must ask before recommending if not already provided by the user. In our demo, mandatory preferences are \textit{location} and \textit{cuisine\_type} as shown in Table \ref{tab:state}, but this selection is easily customized.  Once mandatory preferences have been provided,   
the system will \textit{Answer} if the user has made an inquiry and \textit{Recommend and Explain} otherwise.

\subsection{Retrieval-Augmented Recommendation \emph{and} Explanation} \label{sec:recommendation}
To leverage expressive user review content in \textit{RA-Rec}, we provide a novel 
adaptation of retrieval-augmented generation~\cite{lewis2020rag} for late fusion recommendation and explanation. 
%
To do this, we first generate a query based on semi-structured preferences in the dialogue state and then retrieve relevant items using RIR (Sec. \ref{sec:RIR}) over both the item reviews and known metadata.
This process is illustrated in Figure \ref{fig:late_fusion} with relevant prompts summarized in Table \ref{tab:prompts}. 

Specifically, after a NL query is generated from the hard and soft constraints in the state, we implement late fusion RIR to retrieve a list of top-$k$ scoring items. 
Our implementation of late fusion RIR uses a TAS-B dense encoder \cite{hofstatter2021efficiently} (a variant of BERT \cite{devlin2018bert} fine-tuned for retrieval), dot product similarity, and approximate maximum-inner product search (MIPS) via FAISS \cite{douze2024faiss} to enable scalability to large review corpora. After the top-$k$ items ($k=2$ in our demo) are retrieved, we use the metadata and top-scoring reviews for each item in a prompt to generate a recommendation and explanation of how these items match the dialogue state preferences.

\subsection{Retrieval-Augmented Question Answering} \label{sec:QA}

As observed by Lyu et al. \cite{lyu2021workflow}, the later stages of a recommendation conversational often involve a number of inquiries about the recommended item to confirm that it meets the user's requirements.  To address such QA,  
\textit{RA-Rec} retrieves relevant reviews or metadata for each of the items in question and uses this retrieved information to generate an answer -- with Table \ref{tab:prompts} outlining the prompts used in our QA approach. Our framework is capable of addressing both \emph{individual item questions} such as ``\textit{What kind of menu do they offer?}'' as well as \emph{comparative questions} such as ``\textit{How do their prices compare?}'' as demonstrated in the video (see Sec. \ref{sec:intro}).

In more detail, the first step of QA uses prompting to determine whether an inquiry can be answered using available metadata, which is typically the best knowledge source for simple questions about common properties. In our restaurant recommendation demo, such common metadata fields include price, delivery availability, and parking information. If the inquiry cannot be answered with metadata, a NL query is generated from the user utterance and used to retrieve several reviews for each item in question.  As discussed above, reviews are an expressive knowledge source, especially when inquiries and preferences are stated in complex ways. Finally, the retrieved reviews and metadata for each item are used to generate an answer to the question, which may include item comparisons.  

\subsection{\textit{RA-Rec} System Summary}
In summary, \textit{RA-Rec} employs an LLM-driven, modular DST structure 
to facilitate a controllable recommendation dialogue that can connect complex NL user preferences to matching items using their reviews and metadata. Its JSON semi-structured NL state features configurable keys for domain-specific control while the LLM-updated state values are able to express NL nuance. This state supports novel retrieval-augmented recommendation, explanation, and QA, using scalable retrieval methods such as late fusion RIR and leveraging item reviews and metadata to generate responses.

\section{Demonstration Details}
Our system is designed for easy adaptation to various domains, and as a demonstration, we present \textit{RA-Rec} for restaurant recommendation --- see Sec. \ref{sec:intro} for demo links. Specifically, we use the Yelp Academic Dataset\footnote{\href{https://www.yelp.com/dataset/download}{https://www.yelp.com/dataset/download}} to obtain metadata and over 46K reviews for 1298 restaurants in Edmonton, Alberta.\footnote{The median number of reviews per restaurant was 21.} GPT-3.5-turbo is the LLM used for all prompting steps, but the \textit{RA-Rec} framework is LLM-agnostic and will work with any prompt-based LLM model. 

\section{Future Work}

\textit{RA-Rec} is a flexible LLM prompt-driven architecture and thus opens many new directions for ConvRec systems to support natural user workflows~\cite{lyu2021workflow,jannach2021survey}.  Key extensions include support for (1)~active preference elicitation to narrow down large item spaces \cite{austin2024bayesian}, (2)~structured reasoning over multi-aspect NL preferences \cite{zhang2023recipe}, and (3)~trade-off negotiation between multiple recommendations.

\newpage

\bibliographystyle{ACM-Reference-Format}
\balance
\bibliography{refs}

\end{document}